\documentclass{ecai}

\usepackage{times}
\usepackage{graphicx}
\usepackage{latexsym}

\usepackage{subfigure}
\usepackage{amsmath}
\usepackage{amssymb}
\usepackage{makecell}
\usepackage{multicol}
\usepackage{tablefootnote}
\usepackage[utf8]{inputenc} 
\usepackage[T1]{fontenc}    
\usepackage{hyperref}       
\usepackage{url}            
\usepackage{booktabs}       
\usepackage{amsfonts}       
\usepackage{nicefrac}       
\usepackage{microtype}      
\usepackage{mathtools}
\usepackage{makecell}
\usepackage{arydshln}
\usepackage{algorithm}
\usepackage{algorithmic}
\usepackage{multirow}

\newcommand{\eref}[1]{Eq.~(\ref{#1})}

\newcommand{\cref}[1]{Condition~(\ref{#1})}

\newcommand{\secref}[1]{Section~\ref{#1}}

\let\leq\leqslant
\let\geq\geqslant
\newcommand{\defeq}{\vcentcolon=}

\DeclareMathOperator*{\argmin}{arg\,min}
\DeclareMathOperator{\sign}{sign}
\newcommand{\NAME}[0]{FasTR}

\begin{document}

\title{Fast and Scalable Estimator for Sparse and Unit-Rank Higher-Order Regression Models}

\author{
  Jiaqi Zhang \institute{College of Software Engineering, Southeast University, China} \and Beilun Wang \institute{School of Computer Science and Engineering, Southeast University, China}
}

\maketitle
\bibliographystyle{ecai}

\begin{abstract}
Because tensor data appear more and more frequently in various scientific researches and real-world applications, analyzing the relationship between tensor features and the univariate outcome becomes an elementary task in many fields. To solve this task, we propose \underline{Fa}st \underline{S}parse \underline{T}ensor \underline{R}egression model (FasTR) based on so-called unit-rank CANDECOMP/PARAFAC decomposition. FasTR first decomposes the tensor coefficient into component vectors and then estimates each vector with $\ell_1$ regularized regression. Because of the independence of component vectors, FasTR is able to solve in a parallel way and the time complexity is proved to be superior to previous models. We evaluate the performance of FasTR on several simulated datasets and a real-world fMRI dataset. Experiment results show that, compared with four baseline models, in every case, FasTR can compute a better solution within less time.  
\end{abstract}

\section{Introduction}
Now, higher-order data, which is also called tensor, frequently occur in various scientific and real-world applications. Specifically, in neuroscience, functional magnetic resonance imaging (fMRI) is an example of such tensor data consisting of a series of brain scanning images. Therefore, such data can be characterized as a 3D data (or 3-mode data) with the shape of \emph{time} $\times$ \emph{neuron} $\times$ \emph{neuron}. In many fields, we can encounter the problem that analyzing the relationship between the tensor variable $\mathcal{X}_i \in \mathbb{R}^{\times p_1 \times \cdots \times p_M}$ and the scalar response $y_i$ for every sample $i=1,2,\cdots, N$. Specifically, we assume 
\begin{equation}
    y_i = <\mathcal{W}, \mathcal{X}_i> + \varepsilon_i,
\end{equation}
in which $<\cdot,\cdot>$ is the inner product operator, $\varepsilon_i$ is the noise, and $\mathcal{W} \in \mathbb{R}^{p_1 \times \cdots \times p_M}$ is the coefficient needs to be estimated through regression. Notice that in the real world, these tensor data generally have two properties which makes the coefficient difficult to be inferred perfectly: (1) \textbf{Ultra-high-dimensional setting}, where the number of samples is much less than the number of variables. For example, each sample of the CMU2008 dataset \cite{mitchell2008predicting} is a 3D tensor with shape of $51 \times 61 \times 23$, which is 71553 voxels in total. However, only 360 trials are recorded. The high-dimensional setting will make the estimated solution breaks down because we are trying to infer a large model with a limited number of observations. (2) \textbf{Higher-order structure of data}. The higher-order structure of data exists in many fields, such as fMRI and videos, with the shape of time $\times$ pixel $\times$ pixel. Traditional machine learning methods are proposed for processing vectors or matrices, hence, dealing with high-order data might be a difficulty. In past years, many methods are introduced to address these two problems.

Under high-dimensional settings, several well-known models were already proposed making use of variable selection, such as Lasso \cite{tibshirani1996regression} and Dantzig selector \cite{candes2007dantzig}. Because they are unable to deal with data other than vectors, one naive way to use them on tensor data is vectorization. All the elements in the tensor are stacked into a vector, thence, the existing linear regression can work. However, intuitively, the latent structural information will be lost in such a manner. Therefore, some methods aim at directly handling the tensor. For example, \cite{song2017multilinear} propose Remurs exploiting commonly used $\ell_1$ norm for enforcing sparsity on the estimated coefficient tensor. In addition, a nuclear norm is attached to it to make the solution low-rank. The main shortcoming of Remurs is that the tensor nuclear norm is approximated by the nuclear norm of its unfolding matrices. Because the tensor should be unfolded into matrices, its structure is still destructed. Therefore, though this kind of method is able to obtain an acceptable solution in high-dimensional settings, the higher-order structure is lost.  

To reserve the spatial structure, several methods are introduced based on the CANDECOMP/PARAFAC decomposition, which approximates an M-order tensor through 
\begin{equation}
    \mathcal{A} = \sum\limits_{r=1}^R a^1_r \circ a^2_r \circ \cdots \circ a^M_r.
\end{equation}
Here, $R$ is defined as the CP-rank of the tensor $\mathcal{A}$. For instance, \cite{zhou2013tensor} propose GLTRM which first decomposes the variable tensor and then applies the generalized linear model to estimate each component vector. In addition, \cite{he2018boosted} propose SURF using the divide-and-conquer strategy for each component vector. Almost all the CANDECOMP/PARAFAC-decomposition-based methods, including GLTRM, suffer a problem that the CP-rank $R$ should be pre-specified. However, we always have no prior knowledge about the value of $R$. Even if we can use techniques, such as cross-validation, to estimate $R$ from the data, the solving procedure becomes trivial and computationally expensive for large-scale data. A method called orTRR is previously proposed in \cite{guo2011tensor} automatically obtaining a low-rank coefficient without pre-specifying $R$. But orTRR uses $\ell_2$ norm rather than $\ell_1$ norm for recovering the sparsity of data, which makes it performance poorer than others on variable selection. To our best knowledge, there is no scalable estimator proposed before, for enforcing both sparsity and low-rankness on the solution in high-dimensional settings.

In this paper, we derive ideas of a scalable estimator, Elementary Estimator \cite{yang2014elementary}, and propose \underline{Fa}st \underline{S}parse Higher-Order \underline{T}ensor \underline{R}egression (\NAME{}), which estimates a unit-rank coefficient tensor. First, the problem is decomposed into several sub-problems. Then, for each sub-problem, i.e., each component vector, a closed-form solution can be obtained efficiently. Notice that because the computation of closed-form solution can be speeded up through multi-threading computation or GPUs, thus, the solution of \NAME{} is able to be obtained with small time complexity. See details in \secref{sec:discussion}. To summarize, this paper has the following novelties:

\begin{itemize}
    \item \textbf{A sparse tensor regression model and its fast and scalable solution}: In \secref{sec:method}, we propose a regression model for tensor data, using $\ell_1$ norm for obtaining the sparsity. Moreover, we provide a scalable solution for the model, which is obtained iteratively while at each iteration the temporary estimation has closed form.  
    \item \textbf{State-of-the-art error bound for tensor regression model}: We theoretically prove that our sparse estimator has a state-of-the-art error bound $O((\frac{\sqrt{k_m}\log{p_m}}{Nd_{m,q_M}})^{1/3})$, where $k_m$ denotes the non-zero elements of the $m$th decomposed component of variable tensor and $d_{m,q_m}$ is a constant characterized by the data. Details are shown in \secref{sec:theorem}.
    \item \textbf{Experiments on real-world fMRI dataset}: In \secref{sec:exp}, we make comparison between our \NAME{} and four baselines on several simulated datasets and one fMRI dataset with nine projects. Experiment results empirically show that \NAME{} can obtain better estimations with less time cost.  
\end{itemize}

\section{Notations}
$||\cdot||_1$ denotes the element-wise $\ell_1$ norm, $||\cdot||_*$ denotes the nuclear norm, $||\cdot||_\infty$ denotes the $\ell_\infty$ norm, and $||\cdot||_{\text{spec}}$ denotes the spectral norm. Throughout this paper, the higher-order tensor is denoted by the calligraphic letter $\mathcal{A}$ and the vector is denoted by a lower-case letter $a$. Scalars are also denoted by lower-case letters but stated clearly within the context to avoid confusion.

\section{Background}\label{sec:background}
\subsection{Elmentary estimator for linear regression models (EE-Ridge)}
For vector (first-order) data, \cite{yang2014elementary} propose a state-of-the-art method called EE-Ridge to solve high-dimensional linear regression problems.
Given the sample matrix $X \in \mathbb{R}^{N \times p}$ and the response vector $y \in \mathbb{R}^{N}$, EE-Ridge has the following formulation:
\begin{equation}\label{eq:EE}
  \begin{aligned}
    \widehat{\theta}~=~\argmin\limits_{\theta}&~~~\mathcal{R}(\theta)\\
    \text{s.t.}&~~~\mathcal{R}^*(\theta-(X^TX+\varepsilon I)^{-1}X^Ty) \leq \lambda.
  \end{aligned}
\end{equation}
Here, $\mathcal{R}(u)$ is an arbitrary norm function and $\mathcal{R}^*(u) = \sup_{v:\mathcal{R}(v) \neq 0}\frac{u^Tv}{\mathcal{R}(v)}$. 
The two hyper-parameters, $\varepsilon$ and $\lambda$, handles with the non-invertibility of covariance matrix $X^TX$ and controls the level of sparsity respectively. 
Although EE-Ridge shares certain similarities with the Dantzig selector \cite{candes2007dantzig}, EE-Ridge has outstanding performance on computational complexity.
For instance, when selecting the norm function $\mathcal{R}(\cdot)=||\cdot||_1$, a closed-form solution to \eref{eq:EE} can be obtained through $\widehat{\theta}=S_{\lambda}((X^TX+\varepsilon I)^{-1}X^Ty)$, 
in which $[S_{\lambda}(u)]_i = \sign(u_i)\max(|u_i|-\lambda, 0)$ denotes the soft-thresholding operator. Noticeably, the calculation of this solution is dominated by computing the matrix inversion, which generally acquires $O(p^3)$ time. 
\footnote{Some other methods can compute matrix inversion with lower time complexity. For instance, the Strassen algorithm proposed in \cite{strassen1969gaussian} has the time complexity of $O(p^{2.8})$ on matrix inversion computation.} 
This is a significant improvement in previous variable selectors, such as Lasso and Dantzig selector with the time cost of $O(Tnp^2)$ and $O(p^4)$ respectively. Furthermore, the computation of solution to EE-Ridge can be easily speeded up by the virtue of multiple threads or GPUs.

\subsection{Higher-order tensor regression model}
Given $M$-order predictors $\mathcal{X}_{i} \in \mathbb{R}^{p_1 \times p_2 \times \cdots \times p_M}$ and scalar responses $y_{i}$, $i=1,2,\cdots,N$, higher-order tensor regression models consider that the responses are generated from a linear formulation
$y_{i} = <\mathcal{X}, \mathcal{W}>+\varepsilon_{i}$. Here $\mathcal{W}$ is an $M$-order coefficient tensor and $\varepsilon_i$ is an error term. 
Deriving ideas of LR, the coefficient tensor $\mathcal{W}$ is estimated through
\begin{equation}\label{eq:higher-order-regression}
    \widehat{\mathcal{W}} = \argmin\limits_{\mathcal{W}} ||y-<\mathcal{W}, \mathcal{X}>||_2^2 + \mathcal{R}(\mathcal{W})
\end{equation}
where $\mathcal{R}(\cdot)$ is a norm function enforcing certain properties on the coefficient tensor. 
\eref{eq:higher-order-regression} is akin to the formulation of LR, however, existing LR methods can not be directly applied to it. 
A naive adaption is the vectorization. By stacking the elements of a tensor into a vector first, LR can be utilized. 
However, vectorization will hurt the structural information of data, which makes it inapplicable in real-world applications.     

\subsection{Unit-Rank Tensor CANDECOMP/PARAFAC decomposition}
To reserve latent structural information and decrease the ultra-high dimensionality when dealing with the higher-order tensor , CANDECOMP/PARAFAC decomposition is proposed in \cite{hitchcock1927expression,carroll1970analysis,harshman1970foundations} to decompose the tensor into the outer products of several vectors. Specifically, given a tensor $\mathcal{W} \in \mathbb{R}^{p_1 \times p2 \times \cdots \times p_M}$, it can be decomposed into the outer-products of $M$ component vectors
\begin{equation}\label{eq:CP-decomp}
    \mathcal{W} = w^1 \circ w^2 \circ \cdots \circ w^M,
\end{equation}
in which each $w^m \in \mathbb{R}^{p_m}$. In this way, the number of variables largely reduced from $\prod_{m}p_m$ to $\sum_{m}p_m$. Intuitively, CANDECOMP/PARAFAC decomposition reserves more latent information than simply vectorizing.

\section{Method}\label{sec:method}
  Substituting the M-order coefficient tensor $\mathcal{W} \in \mathbb{R}^{p_1 \times p_M}$ with its CP decomposition $w^1 \circ w^2 \circ \cdots \circ w^M$, the coefficient tensor can be obtained through estimating all the component vectors $w^m,~m=1,2,\cdots,M$. Likewise the linear regression, to infer a certain $w^m$, each sample $\mathcal{X}_i$ needs to be ``projected'' onto the $m$-th space $\mathbb{R}^{p_m}$. Then, intuitively, we aim to let each $w^m$ fit the projected samples corresponding to its space. In addition, instead of coefficient tensor $\mathcal{W}$, we impose sparsity constraints on each component $w^m$. This leads to a more flexible and efficient model because fewer variables need to be dealt with in the high-dimensional settings. Therefore, letting $\mathbf{pr}(\mathcal{X};m) \in \mathbb{R}^{N \times p_m}$ denote the matrix with the $i$-th row bpreing the $\mathcal{X}_i$ projected on the $m$-th space, our objective is solving
  \begin{equation}\label{eq:projected-problem}
      \begin{aligned}
        \widehat{w}^m&=~ \argmin\limits_{w^m \in \mathbb{R}^{p_m}}~~||w^m||_1 \\
        &\text{s.t.}~~||w^m-[\mathbf{pr}^T(\mathcal{X};m)\mathbf{pr}(\mathcal{X};m)+\varepsilon \mathbf{I}]^{-1}\mathbf{pr}^T(\mathcal{X};m)y|| \leq \lambda \\
        &\text{where}~\mathbf{pr}(\cdot~;m): \mathbb{R}^{N \times p_1 \times \cdots \times p_M} \rightarrow \mathbb{R}^{N \times p_m}.
      \end{aligned}
  \end{equation}
  Here, $\lambda$ is a tuning parameter controlling the degree of sparsity and $\mathbf{I} \in \mathbb{R}^{p_m \times p_m}$ is an identity matrix. Parameter $\varepsilon$ aims to make matrix $\mathbf{pr^T_m}(\mathcal{X})\mathbf{pr_m}(\mathcal{X})$ invertible, which handling the crucial problem of high-dimensional learning. Then, fortunately, based on EE-Ridge, \eref{eq:projected-problem} has a closed-form solution
  \begin{equation}\label{eq:projected-solution}
      \begin{aligned}
      \widehat{w}^m ~=~ S_{\lambda}([\mathbf{pr}^T(\mathcal{X};m)\mathbf{pr}(\mathcal{X};m)+\varepsilon \mathbf{I}]^{-1}\mathbf{pr}^T(\mathcal{X};m)y).
      \end{aligned}
  \end{equation}
  With this, as long as $\mathbf{pr}(\mathcal{X};m)$ is easy to be computed, we can solve \eref{eq:projected-problem} directly. 
  
  \subsection{Proposed: \underline{Fa}st \underline{S}parse Higher-Order \underline{T}ensor \underline{R}egression (\NAME{})}
  In our method,  we use an simple and intuitive formulation of the projection function $\mathbf{pr}(\mathcal{X};m)$ as 
  \begin{equation}\label{eq:project-func}
    \begin{aligned}
      \mathbf{pr}(\mathcal{X};m)~=~&\mathcal{X} \times_1 w^1 \times_2 w^2 \times_3 \cdots \times_{m-1} w^{m-1}\\ 
      &~\times_{m+1} w^{m+1} \times_{m+2} \cdots \times_{M} w^M.
     \end{aligned}
  \end{equation}
  Then, substituting $\mathbf{pr}(\mathcal{X};m)$ with \eref{eq:project-func} in \eref{eq:projected-problem}, our \NAME{} aims to solve
  \begin{equation}\label{eq:key-equation}
      \begin{aligned}
        \widehat{w}^m&=~ \argmin\limits_{w^m \in \mathbb{R}^{p_m}}~~||w^m||_1 \\
        &\text{s.t.}~~||w^m-[\mathbf{pr}^T(\mathcal{X};m)\mathbf{pr}(\mathcal{X};m)+\varepsilon \mathbf{I}]^{-1}\mathbf{pr}^T(\mathcal{X};m)y|| \leq \lambda \\
        &\text{where}~~\mathbf{pr}(\mathcal{X}~;m)=\mathcal{X} \times_1 w^1 \times_2 \cdots \times_{m-1} w^{m-1}\\
        &\hspace{1in}\times_{m+1} w^{m+1} \times_{m+2} \cdots \times_{M} w^M
      \end{aligned}
  \end{equation}
  for $m=1,2,\cdots,M$. Notice that the computing of $\mathbf{pr}(\mathcal{X};m)$ is dominated by a large number of multiplications, obtaining the solution \eref{eq:projected-solution} can be easily accelerated by multiple CPUs or GPUs. 
  
  Moreover, we propose a fast algorithm to solve \eref{eq:key-equation} in a component-wise manner. When estimating $w^m$ of a certain $m$, we fix other component vectors as constants. At each iteration, we first compute $\mathbf{pr}(\mathcal{X};m)$ and then get the estimation through \eref{eq:projected-solution}. Specifically, let $w_m^{(t)}$ denotes the estimation of the $m$-th mode component vector at the $t$-th iteration, 
  
  The algorithm is summarized in Algorithm \ref{algorithm}.
  
  \begin{algorithm}[!htb]
    \caption{~~\NAME{}}
    \label{algorithm}
    \begin{algorithmic}[1]
      \STATE {\bfseries Input:} Samples $\mathcal{X} \in \mathbb{R}^{N \times p_1 \times p_2 \times \cdots \times p_M}$, 
      corresponding observed responses $y \in \mathbb{R}^{N}$, , the maximum number of iterations $iter$, and tuning parameters $\lambda$ and $\varepsilon$.  
      \STATE Randomly initialize $w^m,~m=1,2,\cdots,M$
      \FOR{$t=1$ to $\text{iter}$}
        \FOR{$m=1$ to $M$}
          \STATE Compute $\mathbf{pr}(\mathcal{X};m)$ through \eref{eq:project-func};
          \STATE Compute $w^m$ through \eref{eq:projected-solution};
        \ENDFOR
      \ENDFOR
      \STATE $\widehat{\mathcal{W}} = w^1 \circ \cdots \circ w^M$
      \STATE{\bfseries Output:}  $\widehat{\mathcal{W}}$.
    \end{algorithmic}
   \end{algorithm}

\section{Theorem}\label{sec:theorem}
   We now provide a statistical analysis of the component estimator (\eref{eq:key-equation}). 
   We follow the idea of \cite{yang2014elementary} and make following assumptions:
   
   \textbf{(C-Sparse)} The coefficient component $w^m$ is exactly sparse with $k_m$ non-zero elements. 

   \textbf{(C-Ridge)} Let $e_{m,1}, \cdots, e_{m,q_m}, e_{m,q_m+1}, \cdots, e_{m,p_m}$ be the singular vectors 
   of $\frac{1}{N}\mathbf{pr}^T(\mathcal{X};m)\mathbf{pr}(\mathcal{X};m)$ corresponding to the singular 
   values $d_{m,1} \geq \cdots d_{m,q_m} > d_{m,q_m+1} = \cdots = d_{m,p_m} = 0$. Here, $q_m$ is the rank of $\frac{1}{N}\mathbf{pr}^T(\mathcal{X};m)\mathbf{pr}(\mathcal{X};m)$.
   Let $w^m = \sum_{i=1}^{p_m}w^m_ie_{m,i}$. Then, $||\sum_{i=q_m+1}^pw^m_ie_{m,i}||_{\infty} = O(\xi)$ with some equence $\xi \rightarrow 0$.
    
    \begin{theorem}
      Consider assumptions \textbf{(C-Sparse)} and \textbf{(C-Ridge)} are satisfied, 
      there exist positive constants ($c_{m,1}, c_{m,2}$), such that the estimated solution $\widehat{w^m}$ of \eref{eq:key-equation} satisfies
      \begin{equation}\label{eq:error-bound}
        \begin{aligned}
          ||\widehat{w}^m-w^{m*}||_\infty & \leq O((\frac{\sqrt{k_m}\log{p_m}}{Nd_{m,q_m}})^{1/3}) \\
          ||\widehat{w}^m-w^{m*}||_2 & \leq O((\frac{k^2_m\log{p_m}}{Nd_{m,q_m}})^{1/3}) \\
          ||\widehat{w}^m-w^{m*}||_1 & \leq O((\frac{k^3_m\log{p_m}}{Nd_{m,q_m}})^{1/3})
        \end{aligned}
      \end{equation}
      with probability at least $1-c_{m,1}\exp(-c_{m,2}p_m)$. 
      Here, we suppose that $\lambda$ is selected through $\lambda \defeq O((\frac{\sqrt{k_m}\log{p_m}}{Nd_{m,q_m}})^{1/3})$.
    \end{theorem}

\section{Discussion}\label{sec:discussion}
\subsection{Complexity analysis}
When estimating $\widehat{w}^m$ for each mode, the time complexity is dominated by computing $\mathbf{pr}(\mathcal{X};m)$, which costs $O(\prod\limits_{m^\prime \neq m}^{M}p_{m^\prime})$. Once the projection is obtained, $\widehat{w}^m$ is calculated through \eref{eq:projected-solution} with $O(p_m^3)$ time complexity. As the sub-tasks that estimating $w^m$ for each $m$ are independent to each other, these $M$ sub-tasks can be optimized parallelly. Furthermore, notice that the time cost of computing $\widehat{w}^m$ through \eref{eq:projected-solution} can be easily reduced making use of multiple threads or GPUs, this part of time becomes negligible. Therefore, integrating all the ingredients, the time complexity of our method is $O(T \cdot \max\limits_{m}\{\prod\limits_{m^\prime \neq m}^{M}p_{m^\prime}\})$. 

Apart from the computationally efficiency of \NAME{}, our method also acquires small number of memory space. Our method requires two parts of memory space, one of which is used for storing $M$ components $w^m$ and another is for dealing computations on $\mathbf{pr}(\mathcal{X};m)$. Because, among all the computations about the projection, $\mathbf{pr}(\mathcal{X};m)^T\mathbf{pr}(\mathcal{X};m)$ needs the most memory space, the second part requires $O(\max\limits_{m}\{p_m^2\})$ space. Therefore, totally, the memory complexity of \NAME{} is $O(M\max\limits_{m}\{p_m\}+\max\limits_{m}\{p_m^2\})$.

\subsection{Relevance to previous works}
Many methods have been proposed in the literature of regression tasks on higher-order tensor data. In this paper, we focus on the setting that the variables are represented by a tensor $\mathcal{X}$ while the responses are denoted by a vector $\mathcal{y}$. Several models were already recently proposed to estimate the coefficient tensor $\mathcal{W}$ for this specific, what we call, higher-order tensor regression problem. 

One group of these methods is the direct extension of regularized linear regression. Naively, one way to solve this regression problem is vectorization. All the elements in the tensor are first stacked into a vector and then existing linear regression models can be applied to it. One obvious shortcoming of vectorization is that it will cause a loss of latent structural information of the data. To reserve certain potential information, \cite{song2017multilinear} is proposed to estimate a sparse and low-rank coefficient tensor, by integrating the tensor nuclear norm and $\ell_1$ norm into the optimization problem. Notice that in Remurs, the tensor nuclear norm is approximated by the summation of ranks of several unfolded matrices. Remurs still discards some structural information when unfolding the tensor into matrices, although it outperforms than vectorization generally. In addition, \cite{li2019sturm} improve Remurs by substituting the nuclear norm into Tubal nuclear norm \cite{zhang2014novel, zhang2016exact}, which can be efficiently solved through discrete Fourier transform. However, the tensor unfolding is still required. Furthermore, these methods are also computationally expensive because the non-differential regularizer, $\ell_1$ norm or nuclear norm exists in their objective function. Therefore, currently, this group of methods is not a good choice for higher-order tensor regression.

To reserve the latent structure when dealing with tensors, another prevailing group of methods \cite{he2018boosted, zhou2013tensor, guo2011tensor, tan2012logistic} are proposed based on CANDECOMP/PARAFRAC decomposition. Generally, instead of directly estimate the coefficient tensor $\mathcal{W}$, we aim at inferring every component vector $w^m$ in each sub-task. For example, \cite{zhou2013tensor} propose GLTRM using generalized linear model (GLM) to solve each sub-task. Moreover, orTRR is proposed in \cite{guo2011tensor} enforcing sparsity and low-rankness on the estimated coefficient tensor. Instead of $\ell_1$ norm, orTRR utilize $\ell_1$ norm to obtain the sparsity. In addition, recently, \cite{he2018boosted} propose SURF exploiting divide-and-conquer strategy where the sub-task has a similar formulation of Elastic Net \cite{zou2005regularization}. In the paper of SURF, authors empirically show that their method can converge, but a statistical convergence rate is not proved. On the contrary, in this paper, we theoretically prove the error bound of our method. Noticeably, the main limitation of CANDECOMP/PARAFAC-decomposition-based method is that the decomposition rank $R$ should be pre-specified, however, we generally have no prior knowledge about the tensor rank in real-world applications. Although orTRR is able to automatically obtain a low-rank result, the estimated result is sub-optimal due to the fact that $\ell_2$ norm is inferior to $\ell_1$ norm in the sparse setting. Hence, these methods are not suitable for real-world applications.

Some other models were introduced previously for other problem settings. Recently, \cite{hao2019sparse, imaizumi2016doubly, yu2017tensor} propose models for non-parametric estimation by assuming that the response $y_i = f(\mathcal{X}_i) + \text{noise}$, making use of either additive model or Gaussian process. Apart from the above-mentioned ones, many models \cite{sun2017store, raskutti2019convex, yu2016learning, zhu2009intrinsic, li2011multiscale} were put forward to estimate the relationship between the variable tensor $\mathcal{X}$ and a response tensor $\mathcal{Y}$. Another line of statistical models involving tensor data is tensor decomposition \cite{kolda2009tensor, sun2017provable, allen2012sparse, li2017model,madrid2017tensor}. Tensor decomposition can be considered as an unsupervised problem which aims at approximating the tensor with lower-order data. On the contrary, our \NAME{} is a supervised method estimating the latent relationship between variables and responses. Because these methods have different objectives from our method, we pay little attention to them and exclude them from experiments. In section \ref{sec:exp}, we compare \NAME{} with several introduced higher-order tensor regression methods, including Lasso, Elastic Net, Remurs, GLTRM, and SURF.  

\section{Experiment}\label{sec:exp}

\subsection{Experiment Setups}
We experiment on several simulated datasets and a real-world datasets with nine projects to compare the performance of our method with previous methods. We use four previous methods as baselines, which are 1) Linear Regression (Lasso and Elastic Net), 2) Remurs, 3) SURF, and 4) GLTRM. Specifically,  

We use three metrics to evaluate the performance of our method and baselines, including 1) time cost, 2) mean squared error (MSE), and 3) coefficient error (CE). Here, $\text{MSE} = \frac{1}{N}\sum\limits_{i=1}^{N}(\widehat{y}_i-y_i)^2$ and $\text{CE}=\frac{||\widehat{\mathcal{W}}-\mathcal{W}^*||_F}{||\mathcal{W}^*||_F}$.

Furthermore, simulated data are generated through following three steps:
\begin{itemize}
  \item []\textbf{Step 1:}  For $m=1$ to $m$, generate $w^m \in \mathbb{R}^{p_M}$ with each element derived from the Gaussian distribution $\mathcal{N}(0,1)$; 
  given the sparsity degree $s\%$, randomly set $s\%*p_m$ elements of $w^m$ to be $0$; $\mathcal{W} = w^1 \circ \cdots \circ w^M$.
  \item []\textbf{Step 2:}  Generate $\mathcal{X} \in \mathbb{R}^{N \times p_1 \times p_2 \times \cdots \times p_M}$ while each element of $\mathcal{X}$ is 
  derived from the distribution $\mathcal{N}(0,1)$.
  \item []\textbf{Step 3:}  Generate the response $y \in \mathbb{R}^{N}$ with respect to $y = <\mathcal{W},\mathcal{X}>+\alpha\varepsilon$. 
  Here, $\alpha$ controls the degree of noise and each element of $\varepsilon$ is generated from $\mathcal{N}(0,1)$. 
\end{itemize} 
In addtion to simulated datasets, we also use CMU2008 fMRI dataset to show the superiority of our method.

\subsection{Experiments on simulated data}

\begin{figure*}[!tbh]
    \centering
    \subfigure[]{
        \includegraphics[width=\columnwidth]{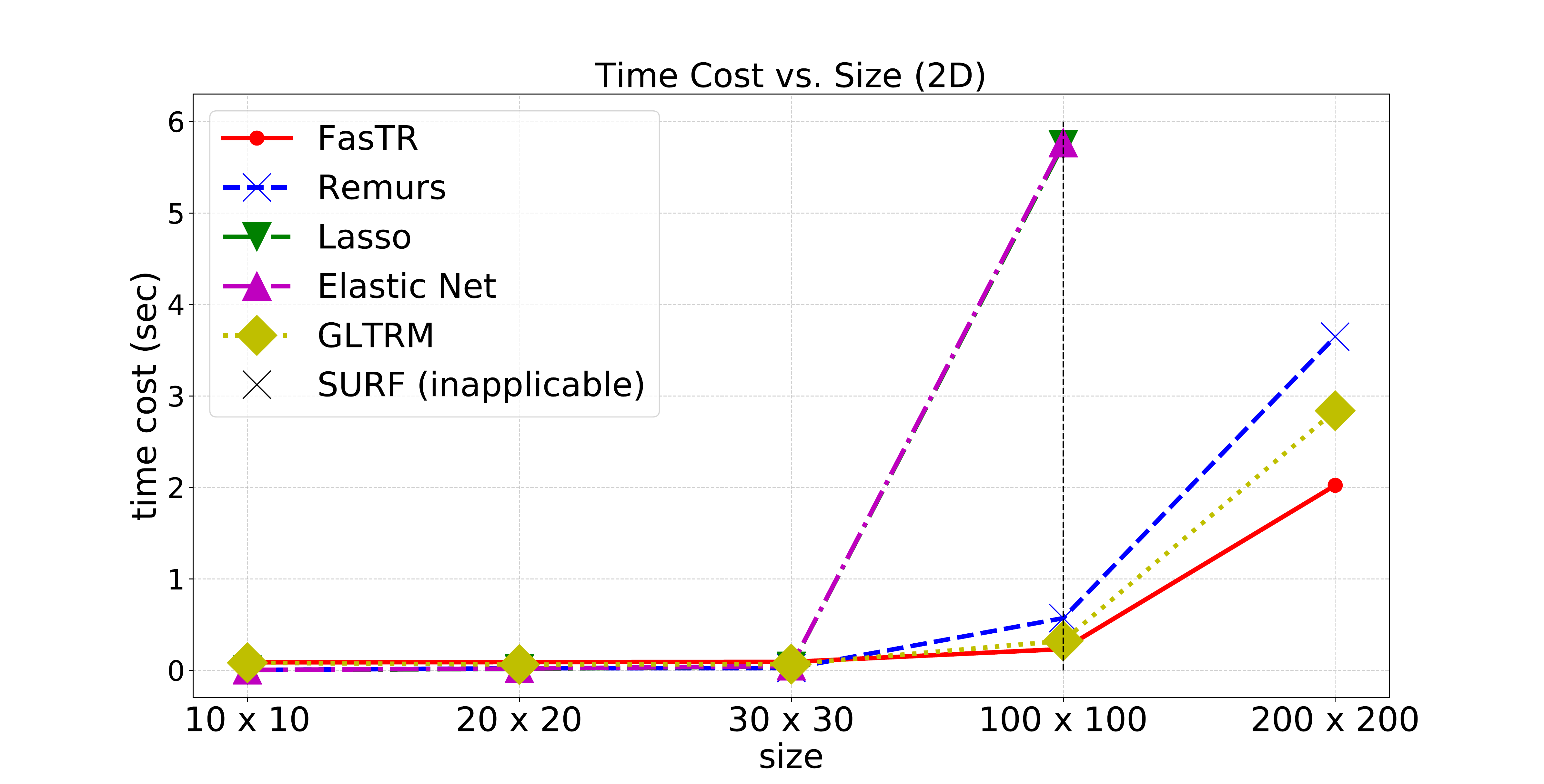}
        \label{fig:timne-2D}
    }
     \subfigure[]{
        \includegraphics[width=\columnwidth]{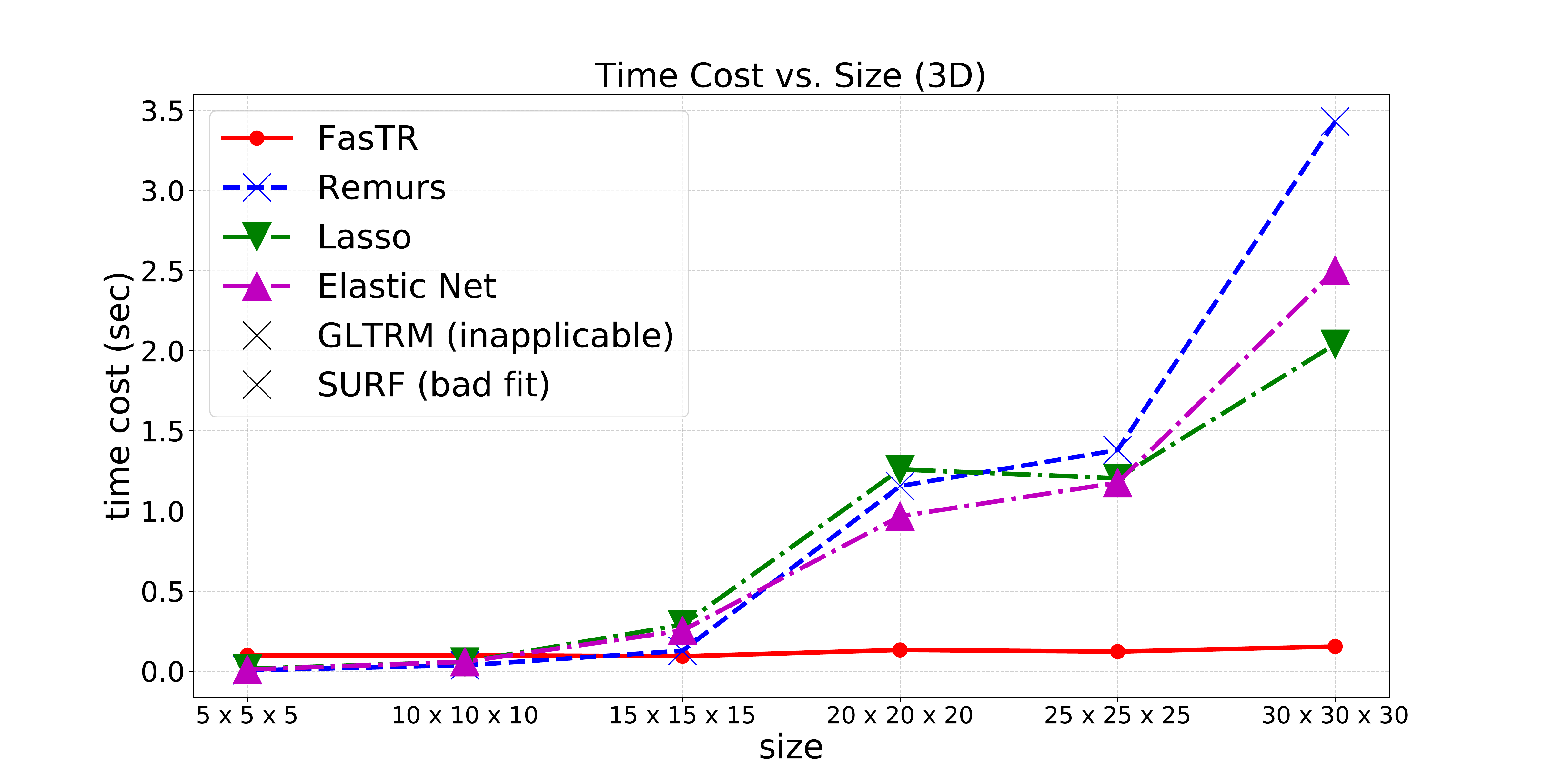}
        \label{fig:timne-3D}
    }
    \caption{(a) Time cost of \NAME{} and baseline methods on simulated 2D data with shape of $10 \times 10$,  $20 \times 20$,  $30 \times 30$,  $100 \times 100$,  and $200 \times 200$. The dashed line indicates that Lasso and Elastic Net are not able to obtain an estimation in 90 seconds. The code of SURF is infeasible for 2D data, hence, its experimental result is omitted. (b) Time cost of \NAME{} and baseline methods on simulated 3D data. with shape of  $5 \times 5 \times 5$, $10 \times 10 \times 10$, $15 \times 15 \times 15$, $20 \times 20 \times 20$, $25 \times 25 \times 25$, and $30 \times 30 \times 30$. GLTRM is only feasible for 2D data, thus, we omit it in the sub-figure. Moreover, the estimation of SURF is terrible, so we also discard it when comparing the time cost.}
    \label{fig:time-cost}
\end{figure*}

When generating simulated datasets, we let the sparsity degree $s\%=20\%$ and noise degree $\alpha=0.1$. Out of fairness, we set the maximal number of iterations to be $1000$ for all the methods and let the method terminate when $\frac{||\mathcal{W}^{t+1}-\mathcal{W}^t||_F}{||\mathcal{W}^t||_F} \leq 1e-3$. 
Moreover, all the tuning parameters of each method are selected through 5-fold cross-validation. The detailed interval of tuning parameters is shown in the appendix. For every single experiment, we run each method 20 times and average the metrics' value over these 20 trials.

To evaluate the superiority of our method, we generate both 2D and 3D datasets varying the data size and the number of samples. For linear regression (LR), we use Lasso and Elastic Net and we report the value of metrics for the method which obtains the better MSE. In Figure \ref{fig:time-cost}, we show the time cost of each method. Notice that because SURF is infeasible for 2D data and GLTRM is infeasible for 3D data, we discard these two methods in the sub-figure correspondingly. Moreover, for 3D data, SURF obtains MSE values much worse that other methods (see Table \ref{tab:accuracy-simulated-data}), thus, we discard SURF in Figure \ref{fig:timne-3D} due to its terrible performance. We can see that our \NAME{} outperforms other baselines and as the dimensionality of data increases, the speeds up of \NAME{} become more and more obvious. Specifically, the two linear regression methods are not able to obtain a solution in less than 90 seconds for data with a shape of $200 \times 200$, while other methods cost no more than 4 seconds. Furthermore, in Table \ref{tab:accuracy-simulated-data} reports the MSE and CE values of every method on each dataset. Noticeably, \NAME{} has better MSE and CE under every setting, compared with baselines, for both 2D and 3D data. Specifically, LR can not compute an estimation because the code throws segmentation fault when the size of data is $300 \times 300 \times 5$ and $400 \times 400 \times 5$, which makes it infeasible for large-scale data. Notice that the performance of linear regression methods are worse than tensor regression methods, which coincide with our statement that the vectorization can do harm to the structural information of data. To summarize, for large-scale data, \NAME{} is able to obtain a better solution with a much less time cost.

\begin{table*}[!htb]
  \centering
  \caption {The MSE and CE of \NAME{} and baselines on several 2D and 3D simulated datasets. SURF is inapplicable for 2D data and GLTRM is inapplicable for 3D data. The LR methods can not obtain the estimation for simulated data with too large size ($300 \times 300 \times 5$ and $400 \times 400 \times 5$) because the MATLAB package ``GLMNET'' throws segmentation fault for too large tensor data. ``LR'' stands for Lasso and Elastic Net and we choose one of them with better MSE to represent linear regression.}
  \label{tab:accuracy-simulated-data}
  \resizebox{\textwidth}{!}{%
  \begin{tabular}{ccccccccccccccc}
  \Xhline{1.2pt}
  \multirow{2}{*}{size (N)}& \multicolumn{2}{c}{\NAME{}} & & \multicolumn{2}{c}{Remurs} & & \multicolumn{2}{c}{LR} & & \multicolumn{2}{c}{SURF} & & \multicolumn{2}{c}{GLTRM}  \\ 
  \cline{2-3} \cline{5-6} \cline{8-9} \cline{11-12} \cline{14-15} 
    & MSE & CE & & MSE & CE & & MSE & CE & & MSE & CE & & MSE & CE \\ \hline
    \multicolumn{15}{c}{\emph{2D Data}}\\
    $10 \times 10~(10)$ & \textbf{0.0427} & \textbf{0.7717} & & 0.1157 & 0.8398 & & 0.7095 & 0.9989 & & \multicolumn{2}{c}{\multirow{6}{*}{\emph{Inapplicable}}} & & 1.8169 & 3.1381 \\
    $20 \times 20~(40)$ & \textbf{0.0305} & \textbf{0.5284} & & 0.2393 & 0.8509 & & 0.4235 & 0.9937 & & & & & 4.5746 & 5.715 \\
    $30 \times 30~(90)$ & \textbf{0.0342} & \textbf{0.3483} & & 1.3414 & 0.8449 & & 0.4980 & 0.9927 & & & & & 7.9937 & 9.2854 \\
    $100 \times 100~(1000)$ & \textbf{0.0628} & \textbf{0.3436} & & 0.1804 & 0.8030 & & 0.5149 & 0.9998 & & & & & 22.4912 & 29.7461 \\
    $200 \times 200~(4000)$ & \textbf{0.056} & \textbf{0.2544} & & 0.4596 & 0.7905 & & 0.5434 & 0.9994 & & & & & 52.3116 & 62.9425 \\
    $300 \times 300~(9000)$ & \textbf{0.0518} & \textbf{0.2349} & & 10.978 & 0.8404 & & 0.5140 & 0.9998 & & & & & 72.3897 & 87.5139 \\
   \cdashline{1-15}[0.8pt/2pt]
   \multicolumn{15}{c}{\emph{3D Data}}\\
  $5 \times 5 \times 5~(100)$ & \textbf{0.0047} & \textbf{0.1205} & & 0.0610 & 0.8524 & & 0.0324 & 0.4507 & & 0.7933 & 0.8417 & & \multicolumn{2}{c}{\multirow{9}{*}{\emph{Inapplicable}}} \\
  $10 \times 10 \times 10~(100)$ & \textbf{0.0127} & \textbf{0.1863} & & 0.0925 & 0.9281 & & 0.0586 & 0.2126 & & 2.8068 & 0.9997 & & \\
  $15 \times 15 \times 15~(100)$ & \textbf{0.1219} & \textbf{0.6772} & & 0.3757 & 0.8707 & & 0.2544 & 0.6461 & & 12.6857 & 0.9989 & & \\ 
  $20 \times 20 \times 20~(100)$ & \textbf{0.2158} & \textbf{0.8323} & & 0.5301 & 0.8562 & & 1.3930 & 1.0711 & & 5.5318 & 0.9619 & & \\ 
  $25 \times 25 \times 25~(100)$ & \textbf{0.1385} & \textbf{0.8820} & & 0.2472 & 0.8929 & & 0.5221 & 1.1443 & & 9.5182 & 0.8948 & & \\
  $30 \times 30 \times 30~(100)$ & \textbf{0.2514} & \textbf{0.8776} & & 0.4386 & 0.8990 & & 1.2765 & 1.0495 & & 21.0606 & 0.9898 & & \\
  $200 \times 200 \times 5~(1000)$ & \textbf{0.3257} & \textbf{0.7334} & & 0.5597 & 0.8877 & & 1.3863 & 1.0010 & & 175.2118 & 0.8436 & & & \\
  $300 \times 300 \times 5~(2250)$ & \textbf{0.1810} & \textbf{0.6901} & & 0.4867 & 0.9116 & & \multicolumn{2}{c}{\multirow{2}{*}{\emph{Inapplicable}}} & & 335.1932 & 0.9988& & \\
  $400 \times 400 \times 5~(4000)$ &  \textbf{0.2105} & \textbf{0.5467} & & 0.4036 & 0.8843 & & & & & 107.6186 & 0.8733 & & \\ 
  \Xhline{1.2pt}
  \end{tabular}
  }
\end{table*}

In high-dimensional settings, we generate a 3D dataset with a shape of $100 \times 100 \times 5$ and vary the number of samples from $300$ to $600$. The sparsity level is set to $s\%=20\%$ and the noise coefficient is fixed to $0.1$. Apparently, Table \ref{tab:high-dimensional} indicates that for every $N$, \NAME{} has much lower MSE value, which indicates that \NAME{} outperform baselines on large-scale and high-dimensional datasets. The MSE value does not reduce along with the increment of the number of samples, which might be a general thought. Because for every $N$, the simulated dataset is generated separately, meaning that these datasets with different $N$ have no relevance. Therefore, in this experiment, the MSE does not have to be improved when more samples are provided. However, in every condition, we show that \NAME{} has better performance. 

\begin{table}
  \centering
  \caption {The MSE value of each method, varying the number of samples. The data size is $100 \times 100 \times 5$ with sparsity level noise coefficient is 0.1 and $s\%=20\%$. GLTRM is inapplicable for 3D data, hence, its MSE value is omitted. ``LR'' stands for Lasso and Elastic Net and we choose one of them with better MSE to represent linear regression.}
  \label{tab:high-dimensional}
  \resizebox{\columnwidth}{!}{%
  \begin{tabular}{cccccc}
  \Xhline{1.2pt}
  \multirow{2}{*}{N}& \multicolumn{5}{c}{MSE Value} \\ 
  \cline{2-6}
   & \NAME{} & Remurs & LR & SURF & GLTRM \\ \hline
  300 & \textbf{0.2529} & 0.4795 & 1.0069 & 22.9196 & \multirow{7}{*}{\emph{Inapplicable}} \\
  350 & \textbf{0.2082} & 0.4451 & 1.0019 & 23.6516 & \\
  400 & \textbf{0.2252} & 0.3692 & 0.8662 & 10.6082 & \\
  450 & \textbf{0.3081} & 0.4902 & 1.2470 & 94.4651 & \\
  500 & \textbf{0.1887} & 0.4896 & 0.8028 & 60.3166 & \\
  550 & \textbf{0.2614} & 0.5480 & 1.0503 & 92.1064 & \\
  600 & \textbf{0.1914} & 0.3370 & 0.7686 & 17.6773 & \\ \Xhline{1.2pt}
  \end{tabular}
  }
\end{table}

\begin{figure*}
    \centering
    \subfigure[]{
        \includegraphics[width=\columnwidth]{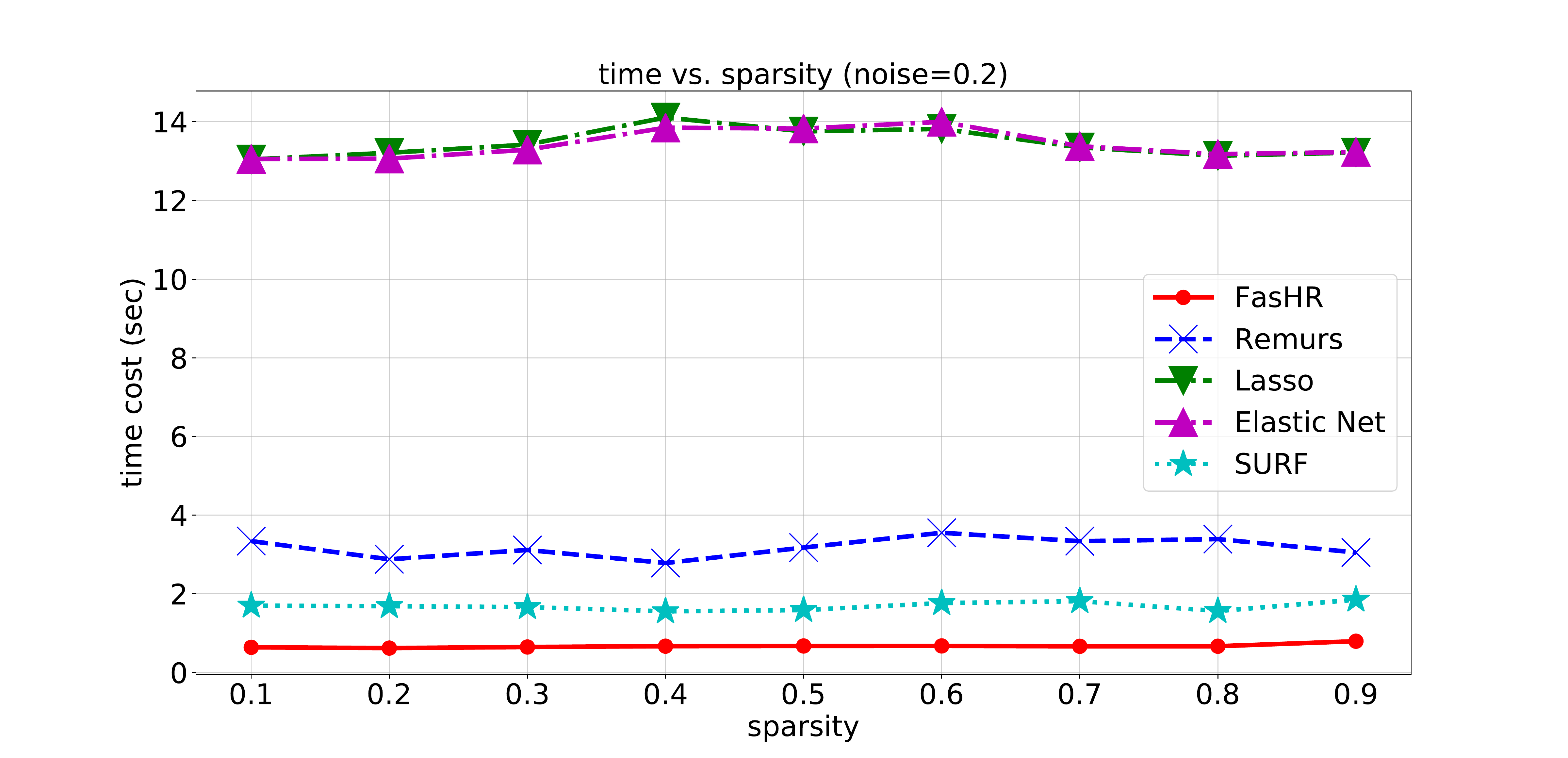}
    }
     \subfigure[]{
        \includegraphics[width=\columnwidth]{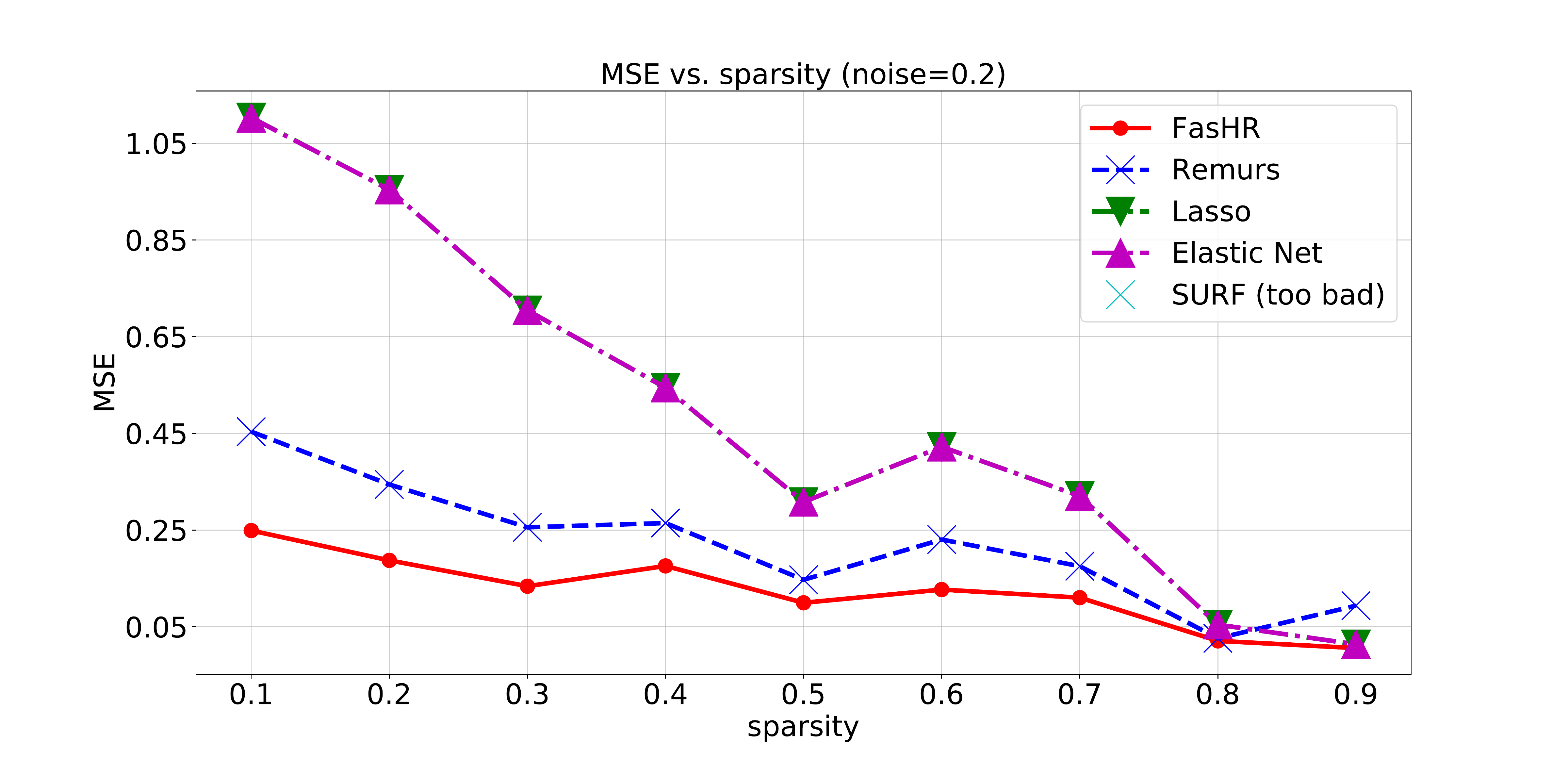}
    }
    \caption{Varying the sparsity level of simulated data, we compare the performance of \NAME{} and baseline methods on 3D data with shape of $100 \times 100 \times 5$ and noise factor of $0.2$. (a) The time cost of methods. We discard GLTRM in this experiment because of its infeasibility for 3D data. (b) The MSE of methods.}
    \label{fig:diff-sparsity}
\end{figure*}

At last, to test the sensitivity of \NAME{} for different sparsity levels, we vary the sparsity level when generating simulated datasets with the shape of $100 \times 100 \times 5$ and the $N = 500$. The noise coefficient is set to 0.2. Varying the sparsity level from $\{10\%, 20\%, 30\%, 40\%, 50\%, 60\%, 70\%, 80\%, 90\%\}$, the time cost and MSE of every method, except GLTRM, are reported in Figure \ref{fig:diff-sparsity}. For different sparsity level, \NAME{} retains the superiority among all the methods. Although under too sparse conditions where the sparsity level is larger than $80\%$, \NAME significantly has lower MSE than others. In a word, \NAME{} obtains better performance on datasets with different level of sparsity, compared with previous methods.

\subsection{Experiments on real-world data}

\begin{table*}[!htb]
  \centering
  \caption {Time cost and AUC value of \NAME{} and baselines on nine classification projects of CMU2008 datasets.}
  \label{tab:CMU2008-accuracy}
  \resizebox{\textwidth}{!}{%
  \begin{tabular}{cccccccccccccccc}
  \Xhline{1.2pt}
  \multirow{2}{*}{Project} & & \multicolumn{2}{c}{\NAME{}} & & \multicolumn{2}{c}{Remurs} & & \multicolumn{2}{c}{Lasso} & & \multicolumn{2}{c}{Elastic Net} & &  \multicolumn{2}{c}{SURF} \\
  \cline{3-4} \cline{6-7} \cline{9-10} \cline{12-13} \cline{15-16}
  & & time & AUC & & time & AUC & & time & AUC & & time & AUC & & time & AUC \\
  \#1 & & 0.3062 & 0.6786 & & 23.9151 & 0.5929 & & 2.2287 & 0.9388 & & 2.6994 & \underline{0.9566} & & 0.1743 & 0.5042 \\
  \#2 & & \textbf{0.3409} & \underline{0.7692} & & 19.2467 & 0.5572 & & 2.6699 & 0.7552 & & 2.6772 & 0.7575 & & 0.1367 & 0.4785\\
  \#3 & & 0.3594 & 0.5314 & & 21.7162 & \underline{0.8642} & & 2.5251 & 0.8109 & & 2.5365 & 0.7786 & & 0.1539 & 0.5286\\
  \#4 & & \textbf{0.3052} & \underline{0.7778} & & 33.0095 & 0.7071 & & 2.4533 & 0.7237 & & 2.4499 & 0.7376 & & 0.1392 & 0.5857\\
  \#5 & & \textbf{0.3073} & \underline{0.7681} & & 26.8979 & 0.7272 & & 2.6739 & 0.5602 & & 2.7003 & 0.5413 & & 0.1285 & 0.4755\\
  \#6 & & \textbf{0.3302} & \underline{0.7222} & & 10.0312 & 0.5972 & & 2.3102 & 0.6554 & & 2.365 & 0.6689 & & 0.1707 & 0.5486\\
  \#7 & & \textbf{0.3453} & \underline{0.6796} & & 19.3765 & 0.5714 & & 2.7856 & 0.6531 & & 2.9838 & 0.679 & & 0.1364 & 0.4444\\
  \#8 & & \textbf{0.2941} & \underline{0.8} & & 16.7164 & 0.5486 & & 2.2362 & 0.6741 & & 2.2854 & 0.7165 & & 0.1811 & 0.5384\\
  \#9 & & \textbf{0.2992} & \underline{0.6953} & & 24.1116 & 0.6428 & & 2.9229 & 0.5531 & & 2.912 & 0.5509 & & 0.196 & 0.5857\\
   \Xhline{1.2pt}
  \end{tabular}
  }
\end{table*}

In this section, we perform fMRI classification tasks on CMU2008 datasets \cite{mitchell2008predicting} with 9 projects in total. Each sample of this dataset is a 3-mode tensor of size $51 \times 61 \times 23$ (71553 voxels).   This classification task aims to predict human activities associated with recognizing the meanings of nouns. Following \cite{kampa2014sparse, song2017multilinear}, we focus on classifications of binary classes: ``tools'' and ``animals''. Here, the class ``tool'' combines observations from ``tool'' and ``furniture'', class ``animal'' combines observations from ``animal'' and ``insect'' in the CMU2008 dataset. Like simulated experiments, values of tuning parameters of each method are selected through 5-fold cross-validation. For each subject, we split the entire dataset into the training dataset and testing dataset with the proportion of $80\%$ and $20\%$ respectively and use AUC to evaluate classification results. Results are shown in Table \ref{tab:CMU2008-accuracy}, which indicates that \NAME{} obtains the best AUC value for most cases. Notice that although SURF has the lowest time cost, the AUC of its solution is around 0.5 and drops below 0.5 sometimes. Hence, we think the classification result of SURF is unacceptable. One interesting result occurs on project \#1, where linear regression methods obtain a much better solution. We think the reason might be that in this subject, the voxels are independent, hence, the data has no latent structure. Anyway, \NAME{} has a significant performance on a real-world fMRI dataset.

\newpage
\bibliography{citations}

\end{document}